\documentclass{article} 
\usepackage{iclr2019_conference,times}
\usepackage{times}
\usepackage{latexsym}
\usepackage{amssymb}
\usepackage{amsthm}
\usepackage{mathtools}
\usepackage{url}
\usepackage{caption}
\usepackage{subcaption}

\usepackage{amsmath,amsfonts,bm}

\def\eqref#1{equation~\ref{#1}}

\def\1{\bm{1}}

\DeclareMathAlphabet{\mathsfit}{\encodingdefault}{\sfdefault}{m}{sl}
\SetMathAlphabet{\mathsfit}{bold}{\encodingdefault}{\sfdefault}{bx}{n}

\usepackage{hyperref}
\usepackage{url}
\usepackage{booktabs}

\newcommand{\ts}{\textsuperscript}

\title{Towards Language Agnostic Universal Representations}

\iclrfinalcopy
\author{Armen Aghajanyan, Xia Song, Saurabh Tiwary \\
Microsoft \\
Bellevue, WA 98004 \\
\texttt{\{araghaja,xiaso,satiwary\}@microsoft.com}
}

\begin{document}

\maketitle

\begin{abstract}
  When a bilingual student learns to solve word problems in math, we expect the student to be able to solve these problem in both languages the student is fluent in, even if the math lessons were only taught in one language. However, current representations in machine learning are language dependent. In this work, we present a method to decouple the language from the problem by learning language agnostic representations and therefore allowing training a model in one language and applying to a different one in a zero shot fashion. We learn these representations by taking inspiration from linguistics and formalizing Universal Grammar as an optimization process \citep{chomsky2014aspects, montague1970universal}. We demonstrate the capabilities of these representations by showing that the models trained on a single language using language agnostic representations achieve very similar accuracies in other languages.
\end{abstract}

\section{Introduction}
Anecdotally speaking, fluent bilingual speakers rarely face trouble translating a task learned in one language to another. For example, a bilingual speaker who is taught a math problem in English will trivially generalize to other known languages. Furthermore there is a large collection of evidence in linguistics arguing that although separate lexicons exist in multilingual speakers the core representations of concepts and theories are shared in memory \citep{altarriba1992representation, mitchel2005bilinguals, bentin1985event}. The fundamental question we're interested in answering is on the learnability of these shared representations within a statistical framework.

We approached this problem from a linguistics perspective. Languages have vastly varying syntactic features and rules. \emph{Linguistic Relativity} studies the impact of these syntactic variations on the formations of concepts and theories \citep{au1983chinese}. Within this framework of study, the two schools of thoughts are linguistic determinism and weak linguistic influence. \emph{Linguistic determinism} argues that language entirely forms the range of cognitive processes, including the creation of various concepts, but is generally agreed to be false \citep{hoijer1954sapir, au1983chinese}. Although there exists some weak linguistic influence, it is by no means fundamental \citep{ahearn2016living}. The superfluous nature of syntactic variations across languages brings forward the argument of \emph{principles and parameters} (PnP) which hypothesizes the existence of a small distributed parameter representation that captures the syntactic variance between languages denoted by parameters (e.g. head-first or head-final syntax), as well as common principles shared across all languages \citep{culicover1997principles}. \emph{Universal Grammar} (UG) is the study of principles and the parameters that are universal across languages \citep{montague1970universal}.

The ability to learn these universalities would allow us to learn representations of language that are fundamentally agnostic of the specific language itself. Doing so would allow us to learn a task in one language and reap the benefits of all other languages without needing multilingual datasets. Our attempt to learn these representations begins by taking inspiration from linguistics and formalizing UG as an optimization problem.

We train downstream models using language agnostic universal representations on a set of tasks and show the ability for the downstream models to generalize to languages that we did not train on.

\section{Related Work}
Our work attempts to unite universal (task agnostic) representations with multilingual (language agnostic) representations \citep{elmo,mccann2017learned}. The recent trend in universal representations has been moving away from context-less unsupervised word embeddings to context-rich representations. Deep contextualized word representations (ELMo) trains an unsupervised language model on a large corpus of data and applies it to a large set of auxiliary tasks \citep{elmo}. These unsupervised representations boosted the performance of models on a wide array of tasks. Along the same lines \cite{mccann2017learned} showed the power of using latent representations of translation models as features across other non-translation tasks. In general, initializing models with pre-trained language models shows promise against the standard initialization with word embeddings. Even further, \cite{radford2017learning} show that an unsupervised language model trained on a large corpus will contain a neuron that strongly correlates with sentiment without ever training on a sentiment task implying that unsupervised language models maybe picking up informative and structured signals.

In the field of multilingual representations, a fair bit of work has been done on multilingual word embeddings. \cite{ammar2016massively} explored the possibility of training massive amounts of word embeddings utilizing either parallel data or bilingual dictionaries via the SkipGram paradigm. Later on an unsupervised approach to multilingual word representations was proposed by \cite{chen2018unsupervised} which utilized an adversarial training regimen to place word embeddings into a shared latent space. Although word embeddings show great utility, they fall behind methods which exploit sentence structure as well as words. Less work has been done on multilingual sentence representations. Most notably both \cite{schwenk2017learning} and \cite{artetxe2017unsupervised} propose a way to learn multilingual sentence representation through a translation task.

We propose learning language agnostic representations through constrained language modeling to capture the power of both multilingual and universal representations. By decoupling language from our representations we can train downstream models on monolingual data and automatically apply the models to other languages.

\section{Universal Grammar as an Optimization Problem}
Statistical language models approximate the probability distribution of a series of words by predicting the next word given a sequence of previous words.
\begin{equation*}
  p(w_0,...,w_n) = \prod_{i=1}^n p(w_i \mid w_0,...,w_{i-1})
\end{equation*}
where $w_i$ are indices representing words in an arbitrary vocabulary.

Learning  grammar is equivalent to language modeling, as the support of $p$ will represent the set of all grammatically correct sentences. Furthermore, let $p_j(\cdot)$ represent the language model for the j\ts{th} language and $w^j$ represents a word from the jth language. Let $k_j$ represent a distributed representation of a specific language along the lines of the PnP argument \citep{culicover1997principles}. UG, through the lens of statistical language modeling, hypothesizes the existence of a factorization of $p_j(\cdot)$ containing a language agnostic segment.
The factorization used throughout this paper is the following:

\begin{align}
  b &= u \circ e_j(w^j_0,...,w^j_i) \label{ug:2}\\
  p_j(w_i \mid w_0,...,w_{i-1}) &= e_j^{-1}(h(b,k_j)) \label{ug:3}\\
  &s.t. \quad d(\mathbf{p}(b\mid j_\alpha) \mid\mid \mathbf{p}(b\mid j_\beta)) \leq \epsilon
\end{align}

\begin{figure}[h]
  \centering
  \includegraphics[height=8.8cm]{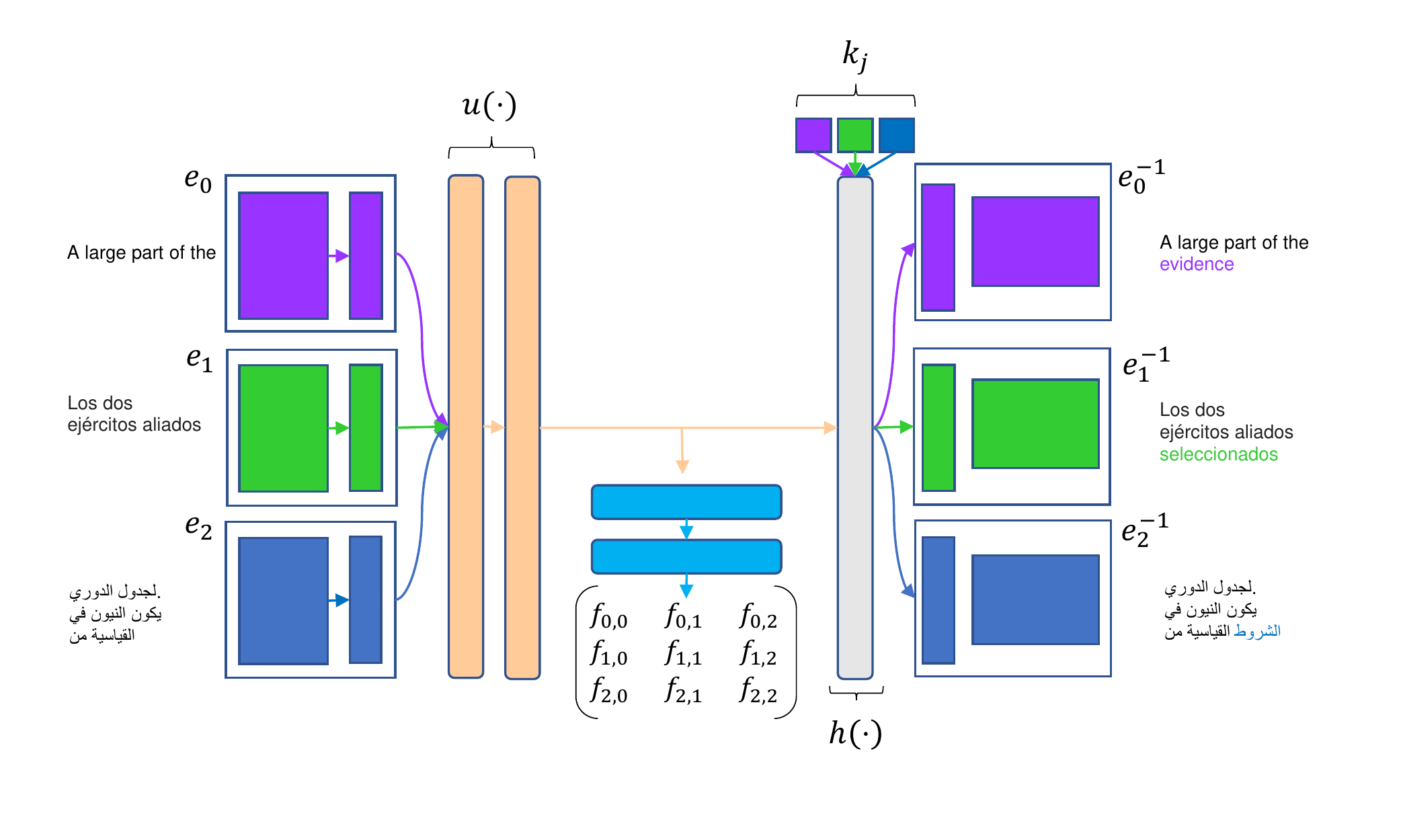}
  \caption{Architecture of UG-WGAN. The amount of languages can be trivially increased by increasing the number of language agnostic segments $k_j$ and $e_j$.}
\end{figure}
The distribution matching constraint $d$, insures that the representations across languages are common as hypothesized by the UG argument.

Function $e_j: \mathbb{N}^i \rightarrow \mathbb{R}^{i \times d}$ is a language specific function which takes an ordered set of integers representing tokens and outputs a vector of size $d$ per token. Function $u: \mathbb{R}^{i \times d} \rightarrow \mathbb{R}^{i \times d}$ takes the language specific representation and attempts to embed into a language agnostic representation. Function $h: (\mathbb{R}^{i \times d}, \mathbb{R}^{f}) \rightarrow \mathbb{R}^{i \times d}$ takes the universal representation as well as a distributed representation of the language of size $f$ and returns a language specific decoded representation. $e^{-1}$ maps our decoded representation back to the token space.

For the purposes of distribution matching we utilize the GAN framework. Following recent successes we use Wasserstein-1 as our distance function $d$ \citep{arjovsky2017wasserstein}.

Given two languages $j_\alpha$ and $j_\beta$ the distribution of the universal representations should be within $\epsilon$ with respect to the $W_1$ of each other. Using the Kantarovich-Rubenstein duality we define

\begin{equation}
  d(\mathbf{p}(b\mid j_\alpha) \mid\mid \mathbf{p}(b\mid j_\beta)) =
   \sup_{||f_{\alpha,\beta}||_L \leq 1} \mathbb{E}_{x\sim\mathbf{p}(b\mid j_\alpha)}\left[f_{\alpha,\beta}(x)\right] - \mathbb{E}_{x\sim\mathbf{p}(b\mid j_\beta)}\left[f_{\alpha,\beta}(x)\right]
\end{equation}
where $L$ is the Lipschitz constant of $f$. Throughout this paper we satisfy the Lipschitz constraint by clamping the parameters to a compact space, as done in the original WGAN paper \citep{arjovsky2017wasserstein}. Therefore the complete loss function for $m$ languages each containing $N$ documents becomes:
\begin{equation}
  \max_{\theta} \sum_{\alpha=0}^m \sum_{i=0}^N \log p_{j_\alpha}(w_{i,0}^\alpha,...,w_{i,n}^\alpha; \theta)\nonumber - \frac{\lambda}{m^2}\sum_{\alpha=0}^m \sum_{\beta=0}^m d(\mathbf{p}(b\mid j_\alpha) \mid\mid \mathbf{p}(b\mid j_\beta))
\end{equation}
$\lambda$ is a scaling factor for the distribution constraint loss.

\section{UG-WGAN}\label{UG-WGAN}
Our specific implementation of this optimization problem we denote as UG-WGAN.
Each function described in the previous section we implement using neural networks. For $e_j$ in equation~\ref{ug:2} we use a language specific embedding table followed by a LSTM \citep{hochreiter1997long}. Function $u$ in equation~\ref{ug:2} is simply stacked LSTM's. Function $h$ in equation~\ref{ug:3} takes input from $u$ as well as a PnP representation of the language via an embedding table. Calculating the real inverse of $e^{-1}$ is non trivial therefore we use another language specific LSTM whose outputs we multiply by the transpose of the embedding table of $e$ to obtain token probabilities. For regularization we utilized dropout and locked dropout where appropriate \citep{gal2016theoretically}.

The critic, adopting the terminology from \cite{arjovsky2017wasserstein}, takes the input from $u$, feeds it through a stacked LSTM, aggregates the hidden states using linear sequence attention as described in DrQA \citep{chen2017reading}. Once we have the aggregated state we map to a $m \times m$ matrix from where we can compute the total Wasserstein loss. A Batch Normalization layer is appended to the end of the critic \citep{ioffe2015batch}. The $\alpha, \beta$th index in the matrix correspond to the function output of $f$ in calculating $W_1(\mathbf{p}(b\mid j_\alpha) \mid\mid \mathbf{p}(b\mid j_\beta))$.

We trained UG-WGAN with a variety of languages depending on the downstream task. For each language we utilized the respective Wikipedia dump. From the wikipedia dump we extract all pages using the wiki2text\footnote{\url{https://github.com/rspeer/wiki2text}} utility and build language specific vocabularies consisting of 16k BPE tokens \citep{sennrich2015neural}. During each batch we sample documents from our set of languages which are approximately the same length. We train our language model via BPTT where the truncation length progressively grows from 15 to 50 throughout training. The critic is updated $10$ times for every update of the language model. We trained each language model for 14 days on a NVidia Titan X. For each language model we would do a sweep over $\lambda$, but in general we have found that $\lambda=0.1$ works sufficiently well for minimizing both perplexity and Wasserstein distance.

\subsection{Exploration}
A couple of interesting questions arise from the described training procedure. Is the distribution matching constraint necessary or will simple joint language model training exhibit the properties we're interested in? Can this optimization process fundamentally learn individual languages grammar while being constrained by a universal channel? What commonalities between languages can we learn and are they informative enough to be exploited?

\begin{figure}
  \centering
  \begin{subfigure}{.5\textwidth}
    \centering
    \includegraphics[height=4.4cm]{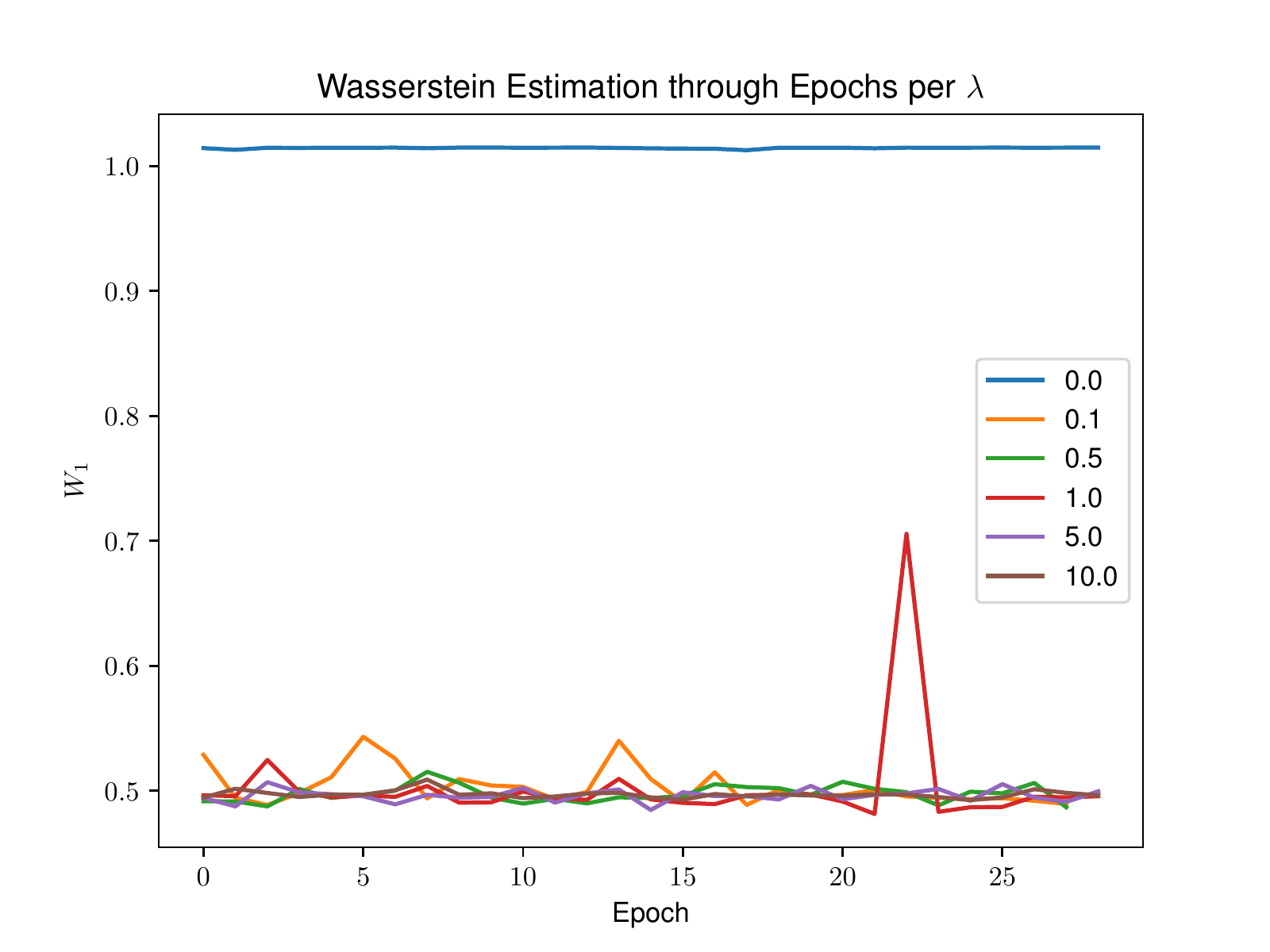}
    \caption{Wasserstein Estimate}
  \end{subfigure}%
  \begin{subfigure}{.5\textwidth}
    \centering
    \includegraphics[height=4.4cm]{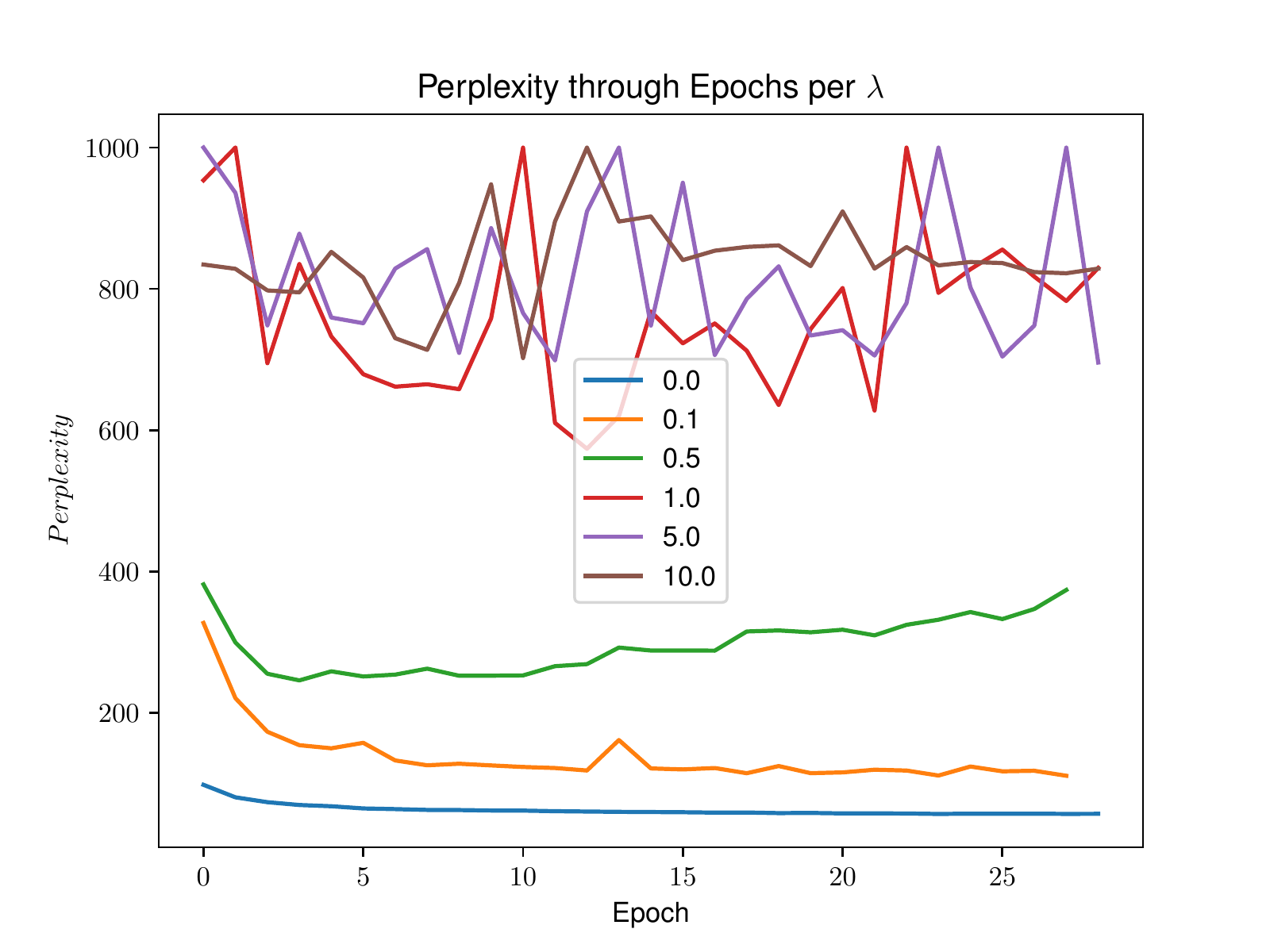}
    \caption{Language Model Perplexity}
  \end{subfigure}
  \caption{Ablation study of $\lambda$. Both Wasserstein and Perplexity estimates were done on a held out test set of documents.}
  \label{fig:ablation}
\end{figure}

\begin{figure}
  \centering
  \begin{subfigure}{.5\textwidth}
    \centering
    \includegraphics[height=4.4cm]{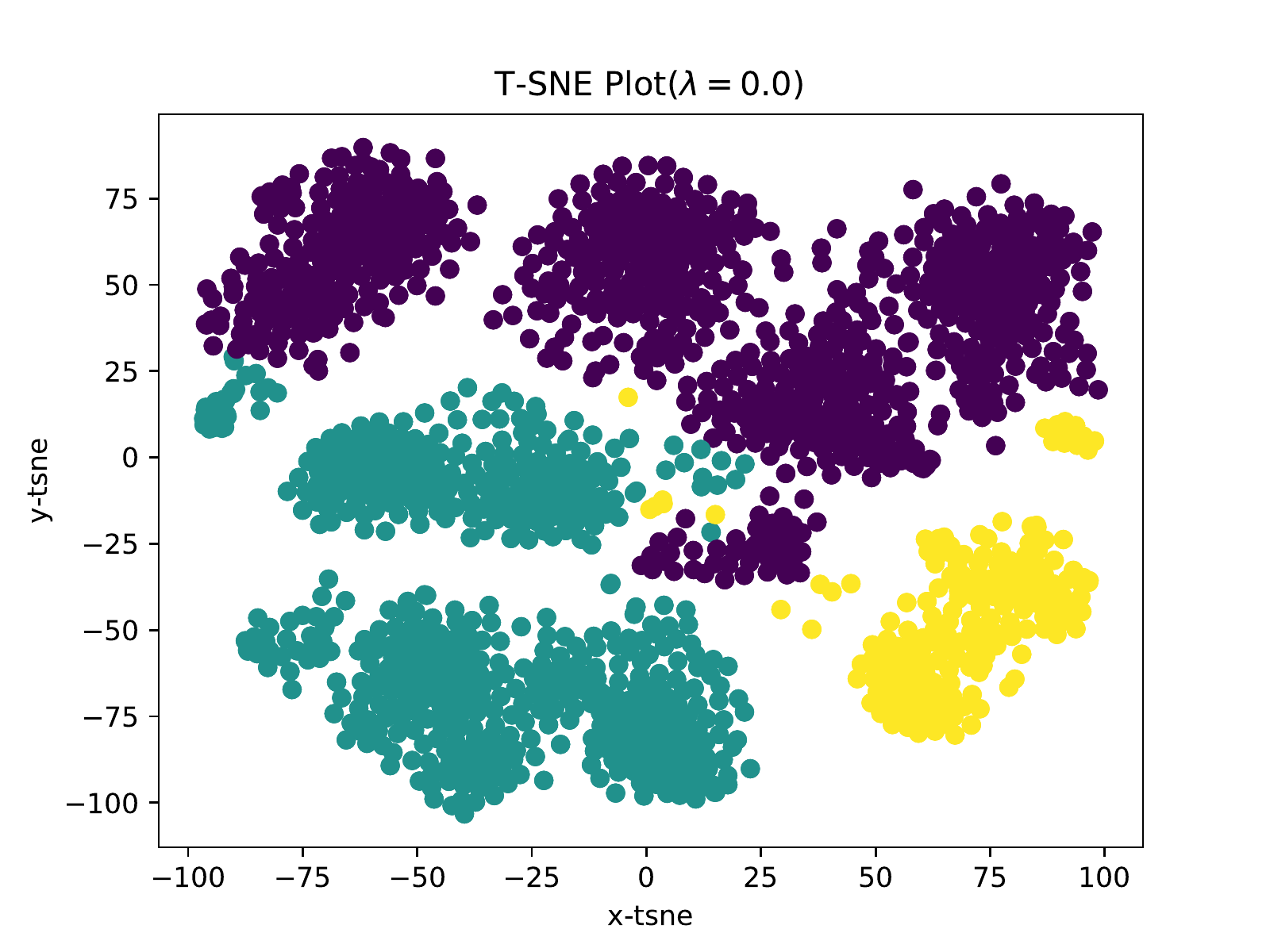}
    \caption{$\lambda=0.0$}
  \end{subfigure}%
  \begin{subfigure}{.5\textwidth}
    \centering
    \includegraphics[height=4.4cm]{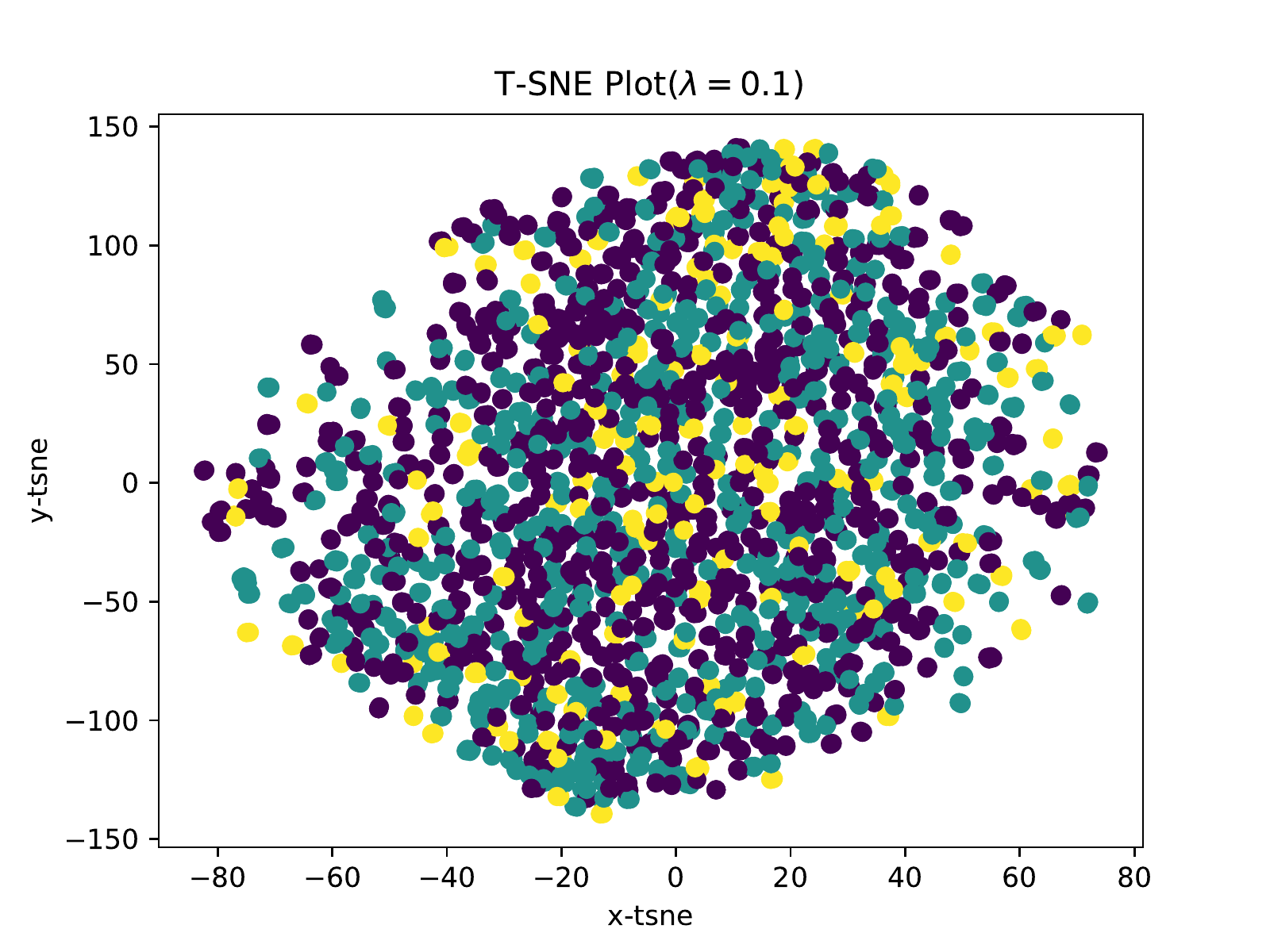}
    \caption{$\lambda=0.1$}
  \end{subfigure}
  \caption{T-SNE Visualization of $u(\cdot)$. Same colored dots represent the same language.}
  \label{fig:tsne}
\end{figure}

We can test out the usefulness of the distribution matching constraint by running an ablation study on the $\lambda$ hyper-parameter. We trained UG-WGAN on English, Spanish and Arabic wikidumps following the procedure described above. We kept all the hyper-parameters consistent apart for augmenting $\lambda$ from 0 to 10. The results are shown in Figure~\ref{fig:ablation}. Without any weight on the distribution matching term the critic trivially learns to separate the various languages and no further training reduces the wasserstein distance. The joint language model internally learns individual language models who are partitioned in the latent space. We can see this by running a t-SNE plot on the universal ($u(\cdot)$) representation of our model and seeing existence of clusters of the same language as we did in Figure~\ref{fig:tsne} \citep{maaten2008visualizing}. An universal model satisfying the distribution matching constrain would mix all languages uniformly within it's latent space.


To test the universality of UG-WGAN representations we will apply them to a set of orthogonal NLP tasks. We will leave the discussion on the learnability of grammar to the Discussion section of this paper.

\section{Experiments}
By introducing a universal channel in our language model we reduced a representations dependence on a single language. Therefore we can utilize an arbitrary set of languages in training an auxiliary task over UG encodings. For example we can train a downstream model only on one languages data and transfer the model trivially to any other language that UG-WGAN was trained on.
\subsection{Sentiment Analysis}
To test this hypothesis we first trained UG-WGAN in English, Chinese and German following the procedure described in Section~\ref{UG-WGAN}. The embedding size of the table was $300$ and the internal LSTM hidden size was 512. A dropout rate of $0.1$ was used and trained with the ADAM optimization method \citep{kingma2014adam}. Since we are interested in the zero-shot capabilities of our representation, we trained our sentiment analysis model only on the english IMDB Large Movie Review dataset and tested it on the chinese ChnSentiCorp dataset and german SB-10K \citep{Sentiment, tan2008empirical}. We binarize the label's for all the datasets.

Our sentiment analysis model ran a bi-directional LSTM on top of fixed UG representations from where we took the last hidden state and computed a logistic regression. This was trained using standard SGD with momentum.

\begin{table}[h]
  \centering
  \begin{tabular}{@{}lrrr@{}}
  \toprule
  Method                     & IMDB    & ChnSentiCorp & SB-10K  \\ \midrule
  NMT + Logistic \citep{schwenk2017learning} & 12.44\%  & 20.12\%      & 22.92\% \\
  FullUnlabeledBow \citep{Sentiment}     & 11.11\% & *           & *      \\
  NB-SVM TRIGRAM \citep{mesnil2014ensemble}        & 8.54\%  & 18.20\%      & 19.40\% \\
  \bf{UG-WGAN} $\lambda=0.1$ + Logistic (Ours) & 8.01\%  & 15.40\%      & 17.32\% \\
  UG-WGAN $\lambda=0.0$ + Logistic (Ours) & 7.80\%  & 53.00\%      & 49.38\% \\
  Sentiment Neuron \cite{radford2017learning}      & 7.70\%  & *           & *      \\
  SA-LSTM \citep{dai2015semi}              & 7.24\%  & *           & *      \\ \bottomrule
  \end{tabular}
  \caption{Zero-shot capability of UG and OpenNMT representation from English training. For all other methods we trained on the available training data. Table shows error of sentiment model.}
\end{table}

We also compare against encodings learned as a by-product of multi-encoder and decoder neural machine translation as a baseline \citep{opennmt}. We see that UG representations are useful in situations when there is a lack of data in an specific language. The language agnostics properties of UG embeddings allows us to do successful zero-shot learning without needing any parallel corpus, furthermore the ability to generalize from language modeling to sentiment attests for the universal properties of these representations. Although we aren't able to improve over the state of the art in a single language we are able to learn a model that does surprisingly well on a set of languages without multilingual data.

\subsection{NLI}
A natural language inference task consists of two sentences; a premise and a hypothesis which are either contradictions, entailments or neutral. Learning a NLI task takes a certain nuanced understanding of language. Therefore it is of interest whether or not UG-WGAN captures the necessary linguistic features. For this task we use the Stanford NLI (sNLI) dataset as our training data in english \citep{bowman2015large}. To test the zero-shot learning capabilities we created a russian sNLI test set by random sampling 400 sNLI test samples and having a native russian speaker translate both premise and hypothesis to russian. The label was kept the same.

For this experiment we trained UG-WGAN on the English and Russian language following the procedure described in Section~\ref{UG-WGAN}. We kept the hyper-parameters equivalent to the Sentiment Analysis experiment. All of the NLI model tested were run over the fixed UG embeddings. We trained two different models from literature,  Densely-Connected Recurrent and Co-Attentive Network by \cite{kim2018semantic} and Multiway Attention Network by \cite{tan2018multiway}. Please refer to this papers for further implementation details.

\begin{table}[h]
  \small
  \centering
  \begin{tabular}{@{}p{10.5cm}rr@{}}
  \toprule
  Method                                                                         & sNLI(en) & sNLI (ru) \\ \midrule
  Densely-Connected Recurrent and Co-Attentive Network Ensemble \citep{kim2018semantic}                 & \bf 9.90\%     & *         \\
  \bf UG-WGAN ($\lambda=0.1$) + Densely-Connected Recurrent and Co-Attentive Network \citep{kim2018semantic}    & 12.25\%  & \bf 21.00\%   \\
  UG-WGAN ($\lambda=0.1$) + Multiway Attention Network  \citep{tan2018multiway}                         & 21.50\%  & 34.25\%   \\
  UG-WGAN ($\lambda=0.0$) + Multiway Attention Network  \citep{tan2018multiway}                           & 13.50\%  & 65.25\%   \\
  UG-WGAN ($\lambda=0.0$) + Densely-Connected Recurrent and Co-Attentive Network \citep{kim2018semantic}  & 11.50\%  & 68.25\%   \\
  Unlexicalized features + Unigram + Bigram features   \citep{bowman2015large}                          & 21.80\%  & 55.00\%   \\ \bottomrule
  \end{tabular}
  \caption{Error in terms of accuracy for the following methods. For \textit{Unlexicalized features + Unigram + Bigram features} we trained on 200 out of the 400 Russian samples and tested on the other 200 as a baseline.}
\end{table}

UG representations contain enough information to non-trivially generalize the NLI task to unseen languages. That being said, we do see a relatively large drop in performance moving across languages which hints that either our calculation of the Wasserstein distance may not be sufficiently accurate or the universal representations are biased toward specific languages or tasks.

One hypothesis might be that as we increase $\lambda$ the cross lingual generalization gap (difference in test error on a task across languages) will vanish. To test this hypothesis we conducted the same experiment where UG-WGAN was trained with a $\lambda$ ranging from $0$ to $10$. From each of the experiments we picked the model epoch which showed the best perplexity. The NLI specific model was the Densely-Connected Recurrent and Co-Attentive Network.

\begin{figure}[h]
  \centering
  \begin{subfigure}{.5\textwidth}
    \centering
    \includegraphics[height=4.4cm]{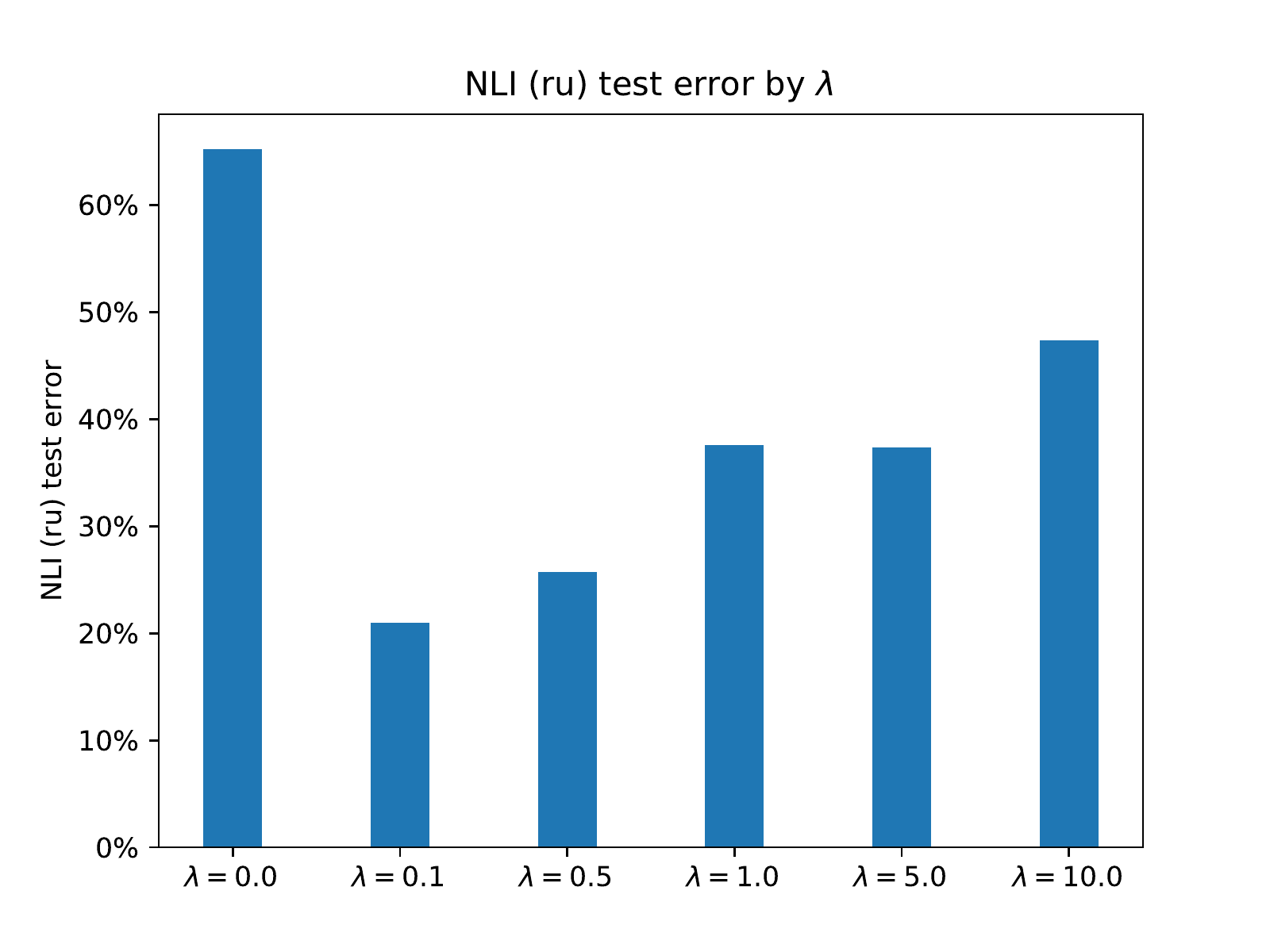}
  \end{subfigure}%
  \begin{subfigure}{.5\textwidth}
    \centering
    \includegraphics[height=4.4cm]{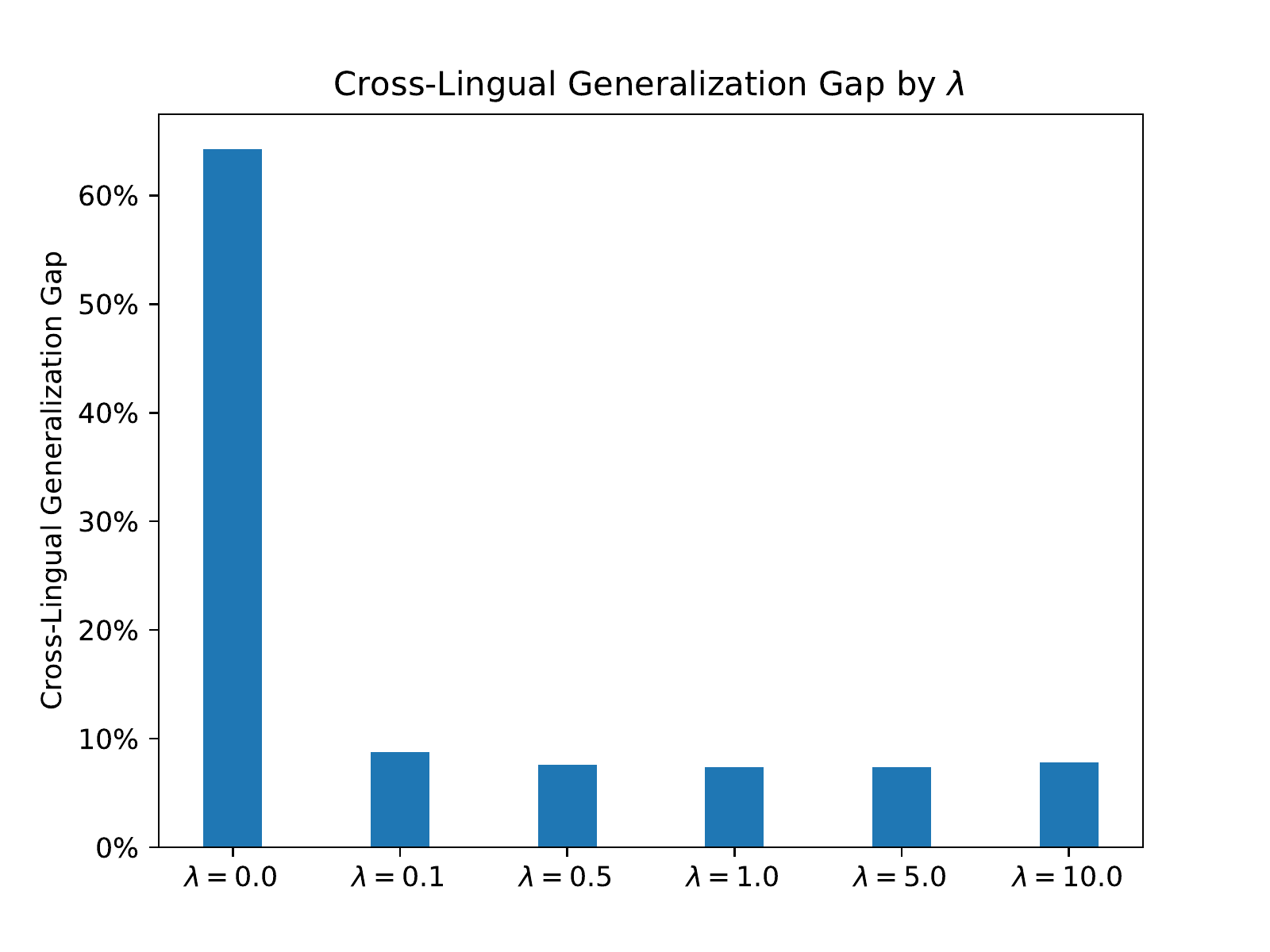}
  \end{subfigure}
  \caption{Cross-Lingual Generalization gap and performance}
  \label{fig:crosslingualgap}
\end{figure}

Increasing $\lambda$ doesn't seem to have a significant impact on the generalization gap but has a large impact on test error. Our hypothesis is that a large $\lambda$ doesn't provide the model with enough freedom to learn useful representations since the optimizations focus would largely be on minimizing the Wasserstein distance, while a small $\lambda$ permits this freedom. One reason we might be seeing this generalization gap might be due to the way we satisfy the Lipschitz constraint. It's been shown that there are better constraints than clipping parameters to a compact space such as a gradient penalty \citep{gulrajani2017improved}. This is a future direction that can be explored.

\section{Discussion}
Universal Grammar also comments on the learnability of grammar, stating that statistical information alone is not enough to learn grammar and some form of native language faculty must exist, sometimes titled the poverty of stimulus (POS) argument \citep{chomsky2010poverty, lewis2001learnability}. From a machine learning perspective, we're interested in extracting informative features and not necessarily a completely grammatical language model. That being said it is of interest to what extent language models capture grammar and furthermore the extent to which models trained toward the universal grammar objective learn grammar.

\begin{figure}[h]
  \centering
  \includegraphics[height=4.4cm]{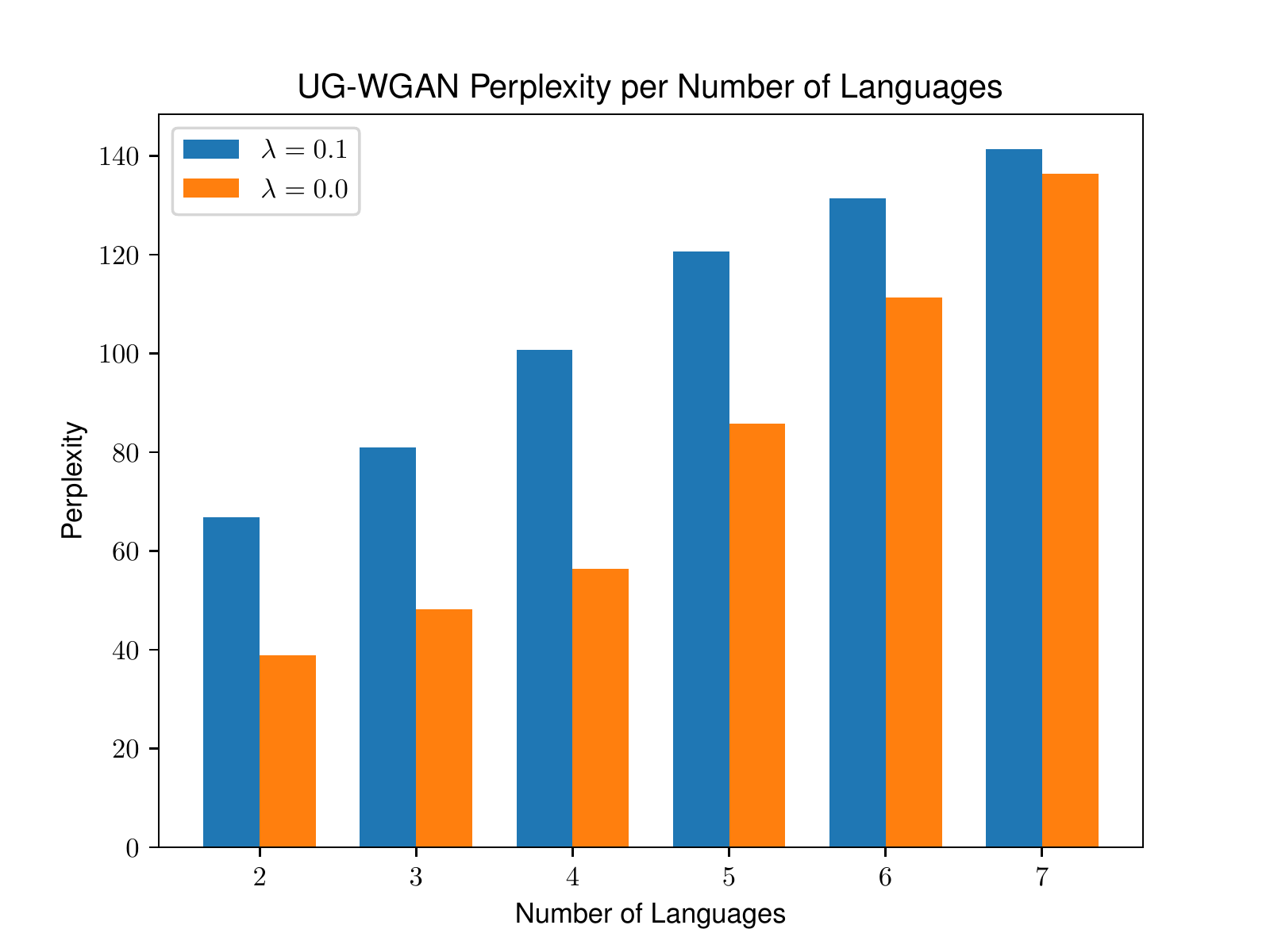}
  \caption{Perplexity calculations on a held out test set for UG-WGAN trained on a varying number of languages.}
  \label{fig:varylanguages}
\end{figure}

One way to measure universality is by studying perplexity of our multi-lingual language model as we increase the number of languages. To do so we trained 6 UG-WGAN models on the following languages: English, Russian, Arabic, Chinese, German, Spanish, French. We maintain the same procedure as described above. The hidden size of the language model was increased to 1024 with 16K BPE tokens being used. The first model was trained on English Russian, second was trained on English Russian Arabic and so on. For arabic we still trained from left to right even though naturally the language is read from right to left. We report the results in Figure~\ref{fig:varylanguages}. As the number of languages increases the gap between a UG-WGAN without any distribution matching and one with diminishes. This implies that the efficiency and representative power of UG-WGAN grows as we increase the number of languages it has to model.

We see from Figure~\ref{fig:ablation} that perplexity worsens proportional to $\lambda$. We explore the differences by sampling sentences from an unconstrained language model and $\lambda=0.1$ language model trained towards English and Spanish in Table~\ref{table:samples}. In general there is a very small difference between a language model trained with a Universal Grammar objective and one without. The Universal Grammar model tends to make more gender mistakes and mistakes due to Plural-Singular Form in Spanish. In English we saw virtually no fundamental differences between the language models. This seems to hint the existence of an universal set of representations for languages, as hypothesized by Universal Grammar. And although completely learning grammar from statistical signals might be improbable, we can still extract useful information.

\begin{table}[t]
  \small
  \begin{tabular}{@{}lp{6.6cm}p{6.6cm}@{}}
  \toprule
   & $\lambda=0.0$ & $\lambda=0.1$ \\ \midrule
  en & earth's oxide is a monopoly that occurs towing of the carbon-booed trunks, resulting in a beam containing of oxygen through the soil, salt, warm waters, and the different proteins. & the practice of epimatic behaviours may be required in many ways of all non-traditional entities. \\\midrule
   & the groove and the products are numeric because they are called "pressibility" (ms) nutrients containing specific different principles that are available from the root of their family, including a wide variety of molecular and biochemical elements. & a state line is a self-government environment for statistical cooperation, which is affected by the monks of canada, the east midland of the united kingdom. \\\midrule
   & however, compared to the listing of special definitions, it has evolved to be congruent with structural introductions, allowing to form the chemical form. & the vernacular concept of physical law is not as an objection (the whis) but as a universal school. \\ \midrule \midrule
  es & la revista más reciente varió el manuscrito originalmente por primera vez en la revista publicada en 1994. & en el municipio real se localiza al mar del norte y su entorno en escajáríos alto, con mayor variedad de cíclica población en forma de cerca de 1070 km2. \\\midrule
   & de hecho la primera canción de "blebe cantas", pahka zanjiwtryinvined cot de entre clases de fanáticas, apareció en el ornitólogo sello triusion, jr., en la famosa publicación playboy de john allen. & fue el último habitantes de suecia, con tres hijos, atasaurus y aminkinano (nuestra). \\\midrule
   & The names of large predators in charlesosaurus include bird turtles hibernated by aerial fighters and ignored fish. & jaime en veracruz fue llamado papa del conde mayor de valdechio, hijo de diego de zúñiga. \\ \bottomrule
  \end{tabular}
  \caption{Example of samples from UG-WGAN with $\lambda=0.0$ and $\lambda=0.1$}
  \label{table:samples}
\end{table}

\section{Conclusion}
In this paper we introduced an unsupervised approach toward learning language agnostic universal representations by formalizing Universal Grammar as an optimization problem. We showed that we can use these representations to learn tasks in one language and automatically transfer them to others with no additional training. Furthermore we studied the importance of the Wasserstein constraint through the $\lambda$ hyper-parameter. And lastly we explored the difference between a standard multi-lingual language model and UG-WGAN by studying the generated outputs of the respective language models as well as the perplexity gap growth with respect to the number of languages.

\bibliography{iclr2019_conference}

\begin{thebibliography}{35}
\providecommand{\natexlab}[1]{#1}
\providecommand{\url}[1]{\texttt{#1}}
\expandafter\ifx\csname urlstyle\endcsname\relax
  \providecommand{\doi}[1]{doi: #1}\else
  \providecommand{\doi}{doi: \begingroup \urlstyle{rm}\Url}\fi

\bibitem[Ahearn(2016)]{ahearn2016living}
Laura~M Ahearn.
\newblock \emph{Living language: An introduction to linguistic anthropology},
  volume~2.
\newblock John Wiley \& Sons, 2016.

\bibitem[Altarriba(1992)]{altarriba1992representation}
Jeanette Altarriba.
\newblock The representation of translation equivalents in bilingual memory.
\newblock \emph{Cognitive processing in bilinguals}, 83:\penalty0 157--174,
  1992.

\bibitem[Ammar et~al.(2016)Ammar, Mulcaire, Tsvetkov, Lample, Dyer, and
  Smith]{ammar2016massively}
Waleed Ammar, George Mulcaire, Yulia Tsvetkov, Guillaume Lample, Chris Dyer,
  and Noah~A Smith.
\newblock Massively multilingual word embeddings.
\newblock \emph{arXiv preprint arXiv:1602.01925}, 2016.

\bibitem[Arjovsky et~al.(2017)Arjovsky, Chintala, and
  Bottou]{arjovsky2017wasserstein}
Martin Arjovsky, Soumith Chintala, and L{\'e}on Bottou.
\newblock Wasserstein gan.
\newblock \emph{arXiv preprint arXiv:1701.07875}, 2017.

\bibitem[Artetxe et~al.(2017)Artetxe, Labaka, Agirre, and
  Cho]{artetxe2017unsupervised}
Mikel Artetxe, Gorka Labaka, Eneko Agirre, and Kyunghyun Cho.
\newblock Unsupervised neural machine translation.
\newblock \emph{arXiv preprint arXiv:1710.11041}, 2017.

\bibitem[Au(1983)]{au1983chinese}
Terry Kit-Fong Au.
\newblock Chinese and english counterfactuals: the sapir-whorf hypothesis
  revisited.
\newblock \emph{Cognition}, 15\penalty0 (1-3):\penalty0 155--187, 1983.

\bibitem[Bentin et~al.(1985)Bentin, McCarthy, and Wood]{bentin1985event}
Shlomo Bentin, Gregory McCarthy, and Charles~C Wood.
\newblock Event-related potentials, lexical decision and semantic priming.
\newblock \emph{Electroencephalography and clinical Neurophysiology},
  60\penalty0 (4):\penalty0 343--355, 1985.

\bibitem[Bowman et~al.(2015)Bowman, Angeli, Potts, and
  Manning]{bowman2015large}
Samuel~R Bowman, Gabor Angeli, Christopher Potts, and Christopher~D Manning.
\newblock A large annotated corpus for learning natural language inference.
\newblock \emph{arXiv preprint arXiv:1508.05326}, 2015.

\bibitem[Chen et~al.(2017)Chen, Fisch, Weston, and Bordes]{chen2017reading}
Danqi Chen, Adam Fisch, Jason Weston, and Antoine Bordes.
\newblock Reading wikipedia to answer open-domain questions.
\newblock \emph{arXiv preprint arXiv:1704.00051}, 2017.

\bibitem[Chen \& Cardie(2018)Chen and Cardie]{chen2018unsupervised}
Xilun Chen and Claire Cardie.
\newblock Unsupervised multilingual word embeddings.
\newblock \emph{arXiv preprint arXiv:1808.08933}, 2018.

\bibitem[Chomsky(2010)]{chomsky2010poverty}
Noam Chomsky.
\newblock Poverty of stimulus: Unfinished business.
\newblock \emph{Transcript of a presentation given at Johannes-Gutenberg
  University, Mainz}, 2010.

\bibitem[Chomsky(2014)]{chomsky2014aspects}
Noam Chomsky.
\newblock \emph{Aspects of the Theory of Syntax}, volume~11.
\newblock MIT press, 2014.

\bibitem[Culicover(1997)]{culicover1997principles}
Peter~W Culicover.
\newblock \emph{Principles and parameters: An introduction to syntactic
  theory}.
\newblock Oxford University Press, 1997.

\bibitem[Dai \& Le(2015)Dai and Le]{dai2015semi}
Andrew~M Dai and Quoc~V Le.
\newblock Semi-supervised sequence learning.
\newblock In \emph{Advances in neural information processing systems}, pp.\
  3079--3087, 2015.

\bibitem[Gal \& Ghahramani(2016)Gal and Ghahramani]{gal2016theoretically}
Yarin Gal and Zoubin Ghahramani.
\newblock A theoretically grounded application of dropout in recurrent neural
  networks.
\newblock In \emph{Advances in neural information processing systems}, pp.\
  1019--1027, 2016.

\bibitem[Gulrajani et~al.(2017)Gulrajani, Ahmed, Arjovsky, Dumoulin, and
  Courville]{gulrajani2017improved}
Ishaan Gulrajani, Faruk Ahmed, Martin Arjovsky, Vincent Dumoulin, and Aaron~C
  Courville.
\newblock Improved training of wasserstein gans.
\newblock In \emph{Advances in Neural Information Processing Systems}, pp.\
  5767--5777, 2017.

\bibitem[Hochreiter \& Schmidhuber(1997)Hochreiter and
  Schmidhuber]{hochreiter1997long}
Sepp Hochreiter and J{\"u}rgen Schmidhuber.
\newblock Long short-term memory.
\newblock \emph{Neural computation}, 9\penalty0 (8):\penalty0 1735--1780, 1997.

\bibitem[Hoijer(1954)]{hoijer1954sapir}
Harry Hoijer.
\newblock The sapir-whorf hypothesis.
\newblock \emph{Language in culture}, pp.\  92--105, 1954.

\bibitem[Ioffe \& Szegedy(2015)Ioffe and Szegedy]{ioffe2015batch}
Sergey Ioffe and Christian Szegedy.
\newblock Batch normalization: Accelerating deep network training by reducing
  internal covariate shift.
\newblock \emph{arXiv preprint arXiv:1502.03167}, 2015.

\bibitem[Kim et~al.(2018)Kim, Hong, Kang, and Kwak]{kim2018semantic}
Seonhoon Kim, Jin-Hyuk Hong, Inho Kang, and Nojun Kwak.
\newblock Semantic sentence matching with densely-connected recurrent and
  co-attentive information.
\newblock \emph{arXiv preprint arXiv:1805.11360}, 2018.

\bibitem[Kingma \& Ba(2014)Kingma and Ba]{kingma2014adam}
Diederik~P Kingma and Jimmy Ba.
\newblock Adam: A method for stochastic optimization.
\newblock \emph{arXiv preprint arXiv:1412.6980}, 2014.

\bibitem[Klein et~al.(2017)Klein, Kim, Deng, Senellart, and Rush]{opennmt}
Guillaume Klein, Yoon Kim, Yuntian Deng, Jean Senellart, and Alexander~M. Rush.
\newblock Open{NMT}: Open-source toolkit for neural machine translation.
\newblock In \emph{Proc. ACL}, 2017.
\newblock \doi{10.18653/v1/P17-4012}.
\newblock URL \url{https://doi.org/10.18653/v1/P17-4012}.

\bibitem[Lewis \& Elman(2001)Lewis and Elman]{lewis2001learnability}
John~D Lewis and Jeffrey~L Elman.
\newblock Learnability and the statistical structure of language: Poverty of
  stimulus arguments revisited.
\newblock In \emph{Proceedings of the 26th Annual Conference on Language
  Development}, 2001.

\bibitem[Maas et~al.(2011)Maas, Daly, Pham, Huang, Ng, and Potts]{Sentiment}
Andrew~L. Maas, Raymond~E. Daly, Peter~T. Pham, Dan Huang, Andrew~Y. Ng, and
  Christopher Potts.
\newblock Learning word vectors for sentiment analysis.
\newblock In \emph{Proceedings of the 49th Annual Meeting of the Association
  for Computational Linguistics: Human Language Technologies}, pp.\  142--150,
  Portland, Oregon, USA, June 2011. Association for Computational Linguistics.
\newblock URL \url{http://www.aclweb.org/anthology/P11-1015}.

\bibitem[Maaten \& Hinton(2008)Maaten and Hinton]{maaten2008visualizing}
Laurens van~der Maaten and Geoffrey Hinton.
\newblock Visualizing data using t-sne.
\newblock \emph{Journal of machine learning research}, 9\penalty0
  (Nov):\penalty0 2579--2605, 2008.

\bibitem[McCann et~al.(2017)McCann, Bradbury, Xiong, and
  Socher]{mccann2017learned}
Bryan McCann, James Bradbury, Caiming Xiong, and Richard Socher.
\newblock Learned in translation: Contextualized word vectors.
\newblock In \emph{Advances in Neural Information Processing Systems}, pp.\
  6294--6305, 2017.

\bibitem[Mesnil et~al.(2014)Mesnil, Mikolov, Ranzato, and
  Bengio]{mesnil2014ensemble}
Gr{\'e}goire Mesnil, Tomas Mikolov, Marc'Aurelio Ranzato, and Yoshua Bengio.
\newblock Ensemble of generative and discriminative techniques for sentiment
  analysis of movie reviews.
\newblock \emph{arXiv preprint arXiv:1412.5335}, 2014.

\bibitem[Mitchel(2005)]{mitchel2005bilinguals}
Aaron Mitchel.
\newblock Do bilinguals access a shared or separate conceptual store? creating
  false memories in a mixed-language paradigm.
\newblock 2005.

\bibitem[Montague(1970)]{montague1970universal}
Richard Montague.
\newblock Universal grammar.
\newblock \emph{Theoria}, 36\penalty0 (3):\penalty0 373--398, 1970.

\bibitem[Peters et~al.(2018)Peters, Neumann, Iyyer, Gardner, Clark, Lee, and
  Zettlemoyer]{elmo}
Matthew~E. Peters, Mark Neumann, Mohit Iyyer, Matt Gardner, Christopher Clark,
  Kenton Lee, and Luke Zettlemoyer.
\newblock Deep contextualized word representations.
\newblock \emph{CoRR}, abs/1802.05365, 2018.
\newblock URL \url{http://arxiv.org/abs/1802.05365}.

\bibitem[Radford et~al.(2017)Radford, Jozefowicz, and
  Sutskever]{radford2017learning}
Alec Radford, Rafal Jozefowicz, and Ilya Sutskever.
\newblock Learning to generate reviews and discovering sentiment.
\newblock \emph{arXiv preprint arXiv:1704.01444}, 2017.

\bibitem[Schwenk \& Douze(2017)Schwenk and Douze]{schwenk2017learning}
Holger Schwenk and Matthijs Douze.
\newblock Learning joint multilingual sentence representations with neural
  machine translation.
\newblock \emph{arXiv preprint arXiv:1704.04154}, 2017.

\bibitem[Sennrich et~al.(2015)Sennrich, Haddow, and Birch]{sennrich2015neural}
Rico Sennrich, Barry Haddow, and Alexandra Birch.
\newblock Neural machine translation of rare words with subword units.
\newblock \emph{arXiv preprint arXiv:1508.07909}, 2015.

\bibitem[Tan et~al.(2018)Tan, Wei, Wang, Lv, and Zhou]{tan2018multiway}
Chuanqi Tan, Furu Wei, Wenhui Wang, Weifeng Lv, and Ming Zhou.
\newblock Multiway attention networks for modeling sentence pairs.
\newblock In \emph{IJCAI}, pp.\  4411--4417, 2018.

\bibitem[Tan \& Zhang(2008)Tan and Zhang]{tan2008empirical}
Songbo Tan and Jin Zhang.
\newblock An empirical study of sentiment analysis for chinese documents.
\newblock \emph{Expert Systems with applications}, 34\penalty0 (4):\penalty0
  2622--2629, 2008.

\end{thebibliography}
\bibliographystyle{iclr2019_conference}

\end{document}